\documentclass[conference]{IEEEtran}
\IEEEoverridecommandlockouts
\usepackage{cite}
\usepackage{amsmath,amssymb,amsfonts}
\usepackage{algorithmic}
\usepackage{graphicx}
\usepackage{textcomp}
\usepackage{xcolor}
\usepackage[pagebackref,breaklinks,colorlinks]{hyperref}
\usepackage{multirow,color,soul,booktabs}
\usepackage{bm}
\usepackage{colortbl}
\definecolor{softred}{rgb}{1.0,0.6,0.6}
\definecolor{softorange}{rgb}{1.0,0.8,0.6}
\definecolor{softyellow}{rgb}{1.0,1.0,0.6}
\usepackage[absolute,overlay]{textpos}
\def\ours{ICG-MVSNet }
\def\oursnospace{ICG-MVSNet}
\def\mdone{Intra-View Fusion }
\def\mdonenospace{Intra-View Fusion}
\def\mdtwo{Cross-View Aggregation }
\def\mdtwonospace{Cross-View Aggregation}

\def\BibTeX{{\rm B\kern-.05em{\sc i\kern-.025em b}\kern-.08em
    T\kern-.1667em\lower.7ex\hbox{E}\kern-.125emX}}
\begin{document}

\title{ICG-MVSNet: Learning Intra-view and Cross-view Relationships for Guidance in Multi-View Stereo}

\author{Yuxi Hu$^{1}$\thanks{This work was supported by the China Scholarship Council (ID: 202208440157). Correspondence to: \texttt{yuxi.hu@tugraz.at}, \texttt{friedrich.fraundorfer@tugraz.at}. Code and supplementary material available at: \url{https://github.com/YuhsiHu/ICG-MVSNet}} \ \ \ Jun Zhang$^{1}$ \ \ \ Zhe Zhang$^{2}$ \ \ \ Rafael Weilharter$^{1}$ \\ Yuchen Rao$^{1}$ \ \ \ Kuangyi Chen$^{1}$ \ \ \ Runze Yuan$^{1}$ \ \ \ Friedrich Fraundorfer$^{1}$ \\
$^{1}$ Graz University of Technology \ \ \ $^{2}$ Peking University \\
}
\maketitle
\begin{abstract}
Multi-view Stereo (MVS) aims to estimate depth and reconstruct 3D point clouds from a series of overlapping images. Recent learning-based MVS frameworks overlook the geometric information embedded in features and correlations, leading to weak cost matching. In this paper, we propose \oursnospace, which explicitly integrates intra-view and cross-view relationships for depth estimation. Specifically, we develop an intra-view feature fusion module that leverages the feature coordinate correlations within a single image to enhance robust cost matching. Additionally, we introduce a lightweight cross-view aggregation module that efficiently utilizes the contextual information from volume correlations to guide regularization. Our method is evaluated on the DTU dataset and Tanks and Temples benchmark, consistently achieving competitive performance against state-of-the-art works, while requiring lower computational resources.
\end{abstract}


\section{Introduction}
\label{sec:intro}

Multi-view Stereo (MVS) is a fundamental area in computer vision that aims to reconstruct 3D geometry from an array of overlapping images. This research domain has evolved significantly over the years, driving progress in autonomous driving and virtual reality. Existing methods typically utilize deep learning to construct cost volumes from multiple camera views and estimate depth maps, thereby simplifying the complex reconstruction task into manageable steps. This depth map-based strategy enhances flexibility and robustness through per-view depth estimation and point cloud fusion.

Recent progress in MVS has featured cascade-based architectures, which adopt a hierarchical approach. Notable examples include~\cite{gu2020cascade, cheng2020deep, yang2020cost, etmvsnet}, which refine predictions progressively from coarse to fine, gradually narrowing down depth hypotheses to optimize computational efficiency. Other strategies, including transformer-based techniques (e.g.,~\cite{ding2022transmvsnet, wang2022mvster, cao2022mvsformer, cao2024mvsformer++}), employ carefully designed external structures to enhance feature extraction. However, they typically miss out on valuable contextual prior and primarily concentrate on pixel-level depth attributes.
Several recent papers have attempted to incorporate geometric information into MVS~\cite{zhang2023geomvsnet, wu2024gomvs}. In particular, GeoMVSNet~\cite{zhang2023geomvsnet} proposes using Gaussian mixture models to represent geometric scene information. However, the feature fusion module they design, which embeds the depth map and raw RGB image into the Feature Pyramid Network (FPN)~\cite{lin2017feature}, results in an overly large and resource-intensive model. 

\begin{figure}[htb]
    \centering
    \includegraphics[width=1\linewidth]{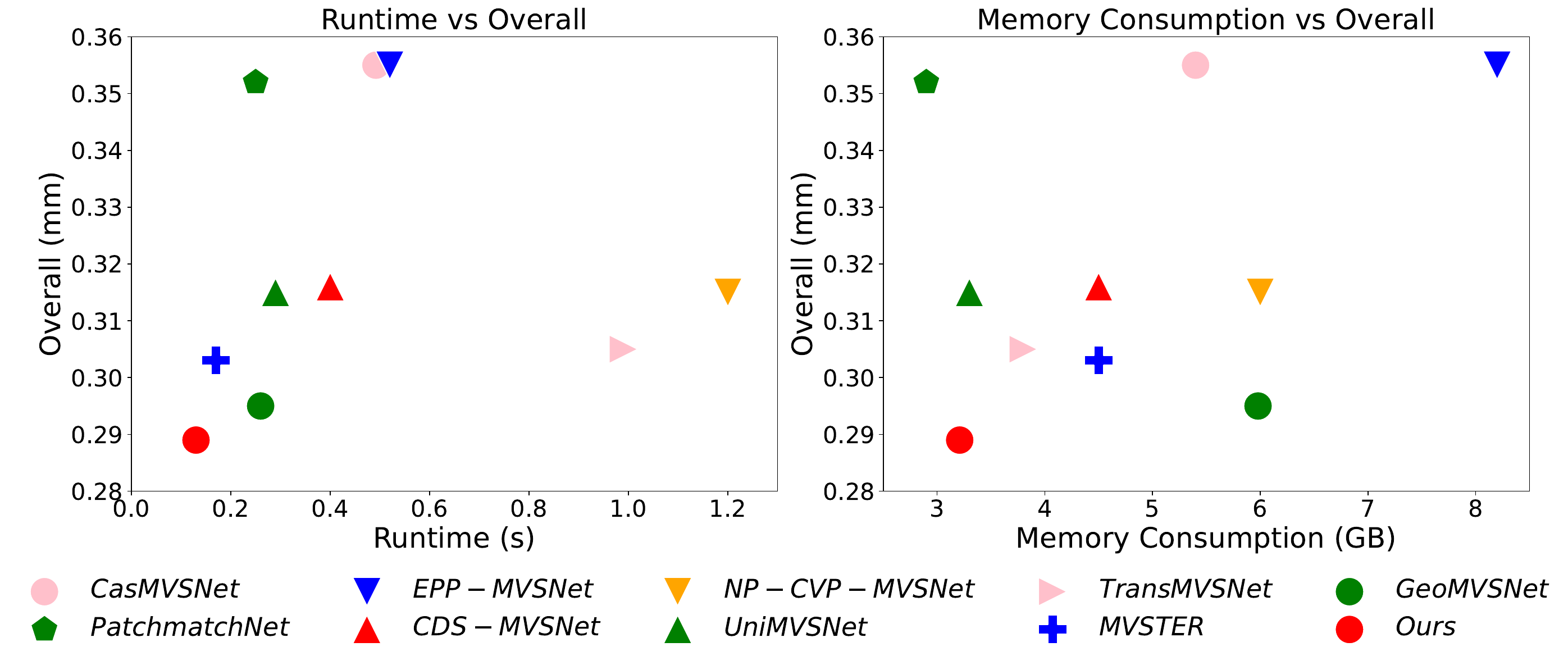}
    \vspace{-20pt}
    \caption{ \textbf{Comparison with state-of-the-art methods in runtime and GPU consumption on DTU~\cite{dtu}.} Our method achieves state-of-the-art performance while maintaining efficient inference time and low memory usage.}
    \label{fig:runtime}
    \vspace{-10pt}
\end{figure}

Diverging from current approaches, we propose investigating information embedded in single-view features and cross-view correlations. Specifically, we leverage positional information in FPN~\cite{lin2017feature} and capture dependencies in one coordinate direction while retaining positional information in the other. Meanwhile, coarse cross-view correlations with abundant geometric information from different views are exploited to guide the correlation distributions of fine stages. We found that pair-wise correlation within cost volumes has rich information from multiple perspectives. Connections exist not only between feature channels under the same depth hypothesis and the same feature channel under different depth hypotheses, but also between different channels at different depth hypotheses. Therefore, we design a cross-view aggregation scheme to directly process correlations across stages, depth hypotheses, and features, resulting in a lightweight but robust cost matching.

Our main contributions are summarized as follows.
\begin{itemize}
    \item We propose \ours that includes \mdonenospace~(IVF), which embeds positional information along two coordinate directions into feature maps within a single image for robust matching,
    \item and \mdtwonospace~(CVA), a lightweight cross-view aggregation scheme that efficiently utilizes the prior guidance from previous correlations.
    \item We conduct extensive experimental comparisons against state-of-the-art methods. The results demonstrate that the proposed method achieves competitive performance with an optimal balance between effectiveness and computational efficiency, as shown in Fig.~\ref{fig:runtime}.
\end{itemize}
\section{Method}
\label{sec:method}

\begin{figure*}[htb]
    \centering
    \includegraphics[width=0.98\linewidth]{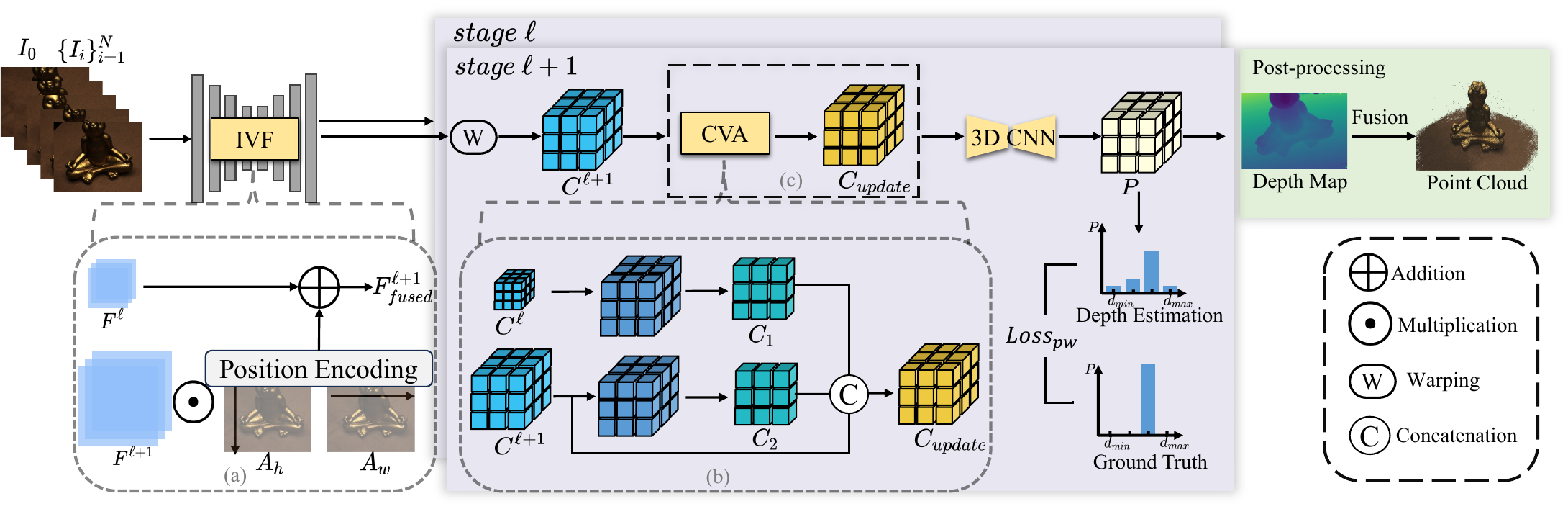}
    \vspace{-10pt}
    \caption{\textbf{The overall architecture.} Our method is a coarse-to-fine framework that estimates depths from low resolution ($stage \ \ell$) to high resolution ($stage \ \ell+1$), where $\ell = 0, 1, 2$, resulting in a total of $4$ stages. Features of reference and source images $\{\bm{F}_{i}\}_{i=0}^{N}$ are extracted by a feature pyramid network with the help of~\mdonenospace~(\textbf{IVF}), whose details are illustrated in (a). The source image features are warped into the $D$ frustum planes of the reference camera and an element-wise multiplication is used to correlate each source image with the reference image. These correlations are aggregated into a single cost volume $\bm{C}$. In finer stages (stage $1$, $2$, and $3$), both current and previous stage correlations are used in~\mdtwonospace~(\textbf{CVA}), whereas in stage $0$, the cost volume is not updated due to the absence of contextual correlations from a previous stage. Details of this process are illustrated in (b) and (c). Regularization (3D CNN) yields the probability volume $\bm{P}$, from which the depth hypothesis with the highest probability is selected for the final depth map. Depth maps from multiple viewpoints are fused into a point cloud, in a non-learnable process.}
    \label{fig:architecture}
\vspace{-15pt}
\end{figure*}

Given a set of images, $\bm{I}_{0}$ denotes the reference image for which the depth is to be estimated, while $\{\bm{I}_{i}\}_{i=1}^{N}$ represents $N$ source images—adjacent images which serve as an auxiliary input for estimating the depth of reference view. Our network estimates the depth map with width $W$ and height $H$.

We employ a coarse-to-fine network for depth estimation. For each pixel, we uniformly sample $D$ depth discrete values within the range defined by the minimum and maximum depths $[d_{min},d_{max}]$. The depth hypothesis $d$ with the highest probability is selected and also used as the center for the next stage. A narrower range of depth hypotheses is then generated around this center, enabling progressively finer depth estimation that converges toward the true depth value. For the number of depth hypothesis planes, we set $D$ to $8,8,4,4$ for each level of the stage $\ell$, ensuring an appropriate balance between precision and computational efficiency.

The overall architecture of our network is illustrated in Fig.~\ref{fig:architecture}. Image features $\{\bm{F}_{i}\}_{i=0}^{N}$ of both reference and source images are first extracted by the FPN~\cite{lin2017feature}. Then the features of source views are warped into the $D$ fronto-parallel planes of the reference camera frustum, denoted as $\{\bm{V}_{i}\}_{i=1}^{N}$. The feature volume of the reference view $\bm{V}_{0}$ is obtained by expanding reference feature $\bm{F}_{0}$ to $D$ depth hypothesis. The element-wise multiplication is performed to obtain the correlation between each warped source volume $\bm{V}_{i}$ and reference volume $\bm{V}_{0}$. After that, multiple feature correlation volumes are aggregated into a single cost volume $\bm{C} \in \mathbb{R}^{G \times D \times H \times W}$, where $G$ is the group-wise correlation channel~\cite{guo2019group}. In the finer stages, the cross-view aggregation module will use both the correlation of the current stage and the previous stage. Afterward, the 3D CNN is applied to obtain the probability volume $\bm{P} \in \mathbb{R}^{D \times H \times W}$, which will be used to select the depth hypothesis with the largest probability in each pixel to obtain the final depth map. With the depth map, the 3D point cloud is generated via fusion.

\subsection{Feature Extraction}
\label{sec:feature}

Existing works~\cite{yao2018mvsnet,yao2019recurrent} extract deep features $\{\bm{F}_i\}_{i=0}^N$ from input images $\{\bm{I}_i\}_{i=0}^N$ using FPN~\cite{lin2017feature}, which do not fully explore the intra-view knowledge.  
GeoMVSNet~\cite{zhang2023geomvsnet} highlights the value of pixel coordinates while integrating depth and RGB images for feature extraction, but this increases GPU usage and runtime due to the added channels and separate convolutions. Moreover, GeoMVSNet~\cite{zhang2023geomvsnet} assumes that the absolute values of coordinates contribute to feature extraction, but this assumption may lack generalization. We focus on determining the relative importance of positions, which is sufficient for capturing coordinate information. Unlike previous methods that rely on Transformer or pre-trained models~\cite{cao2022mvsformer,ding2022transmvsnet}, we exploit intra-view relationships by multiplying features with two attention maps that capture long-range dependencies.

\noindent \textbf{\mdone(IVF).} 
The \mdonenospace~block is a lightweight unit that aims to enhance the expressive ability of the features, which stands out for its simplicity and efficiency. It can encode both long-range dependencies and feature channel relationships with positional information to capture valuable information in images. 
For each feature $\bm{F}_{i}$ in features $\{\bm{F}_{i}\}_{i=0}^{N}$, we omit the index $i$ and denote the pyramid level by $\ell$. At level $\ell$, the feature $\bm{F}^{\ell}$ is in lower resolution, while at level $\ell + 1$, the feature $\bm{F}^{\ell + 1}$ is twice the height $H$ and width $W$. We first calculate the weights as $\bm{T}_{h}$ and $\bm{T}_{w}$ using Eqn.~\ref{equ:attention}. For a feature $x$ in the $c$-th channel, the pooling operation at height $h$ and width $w$ is calculated as follows:
\begin{equation}
    \bm{T}_{h}(h) = \frac{1}{W} \sum_{i=0}^{W} x(h,i) \ ,
    \bm{T}_{w}(w) = \frac{1}{H} \sum_{j=0}^{H} x(j,w) \ .
    \label{equ:attention}
\end{equation}

Following this, the weights $\bm{T}_{h}(h)$ and $\bm{T}_{w}(w)$ are concatenated and passed through a convolution layer for a coordinate fusion which integrates the height and width importance into a unified representation. Then the representation is split again to produce two separate attention maps $\bm{A}_{h}$ and $\bm{A}_{w}$ which encode dependencies along one coordinate direction and preserve positional information along the other direction. Next, the attention maps $\bm{A}_{h}$ and $\bm{A}_{w}$ are broadcasted to match the shape of $\bm{F}^{\ell + 1}$ and applied to the feature $\bm{F}^{\ell + 1}$ using the pixel-wise multiplication. Subsequently, the feature $\bm{F}^\ell$ is upsampled and merged through addition. The fused intra-view feature in the finer stage $\bm{F}^{\ell + 1}_{fused}$ is formulated as:
\begin{equation}
     \bm{F}^{\ell + 1}_{fused} = \bm{F}^{\ell}_{\uparrow} \oplus \bm{F}^{\ell + 1}_{A} \ ,
     \label{equ:branch}
\end{equation}
\begin{equation}
    \bm{F}^{\ell + 1}_{A} = \bm{A}_{h} \odot \bm{A}_{w} \odot \bm{F}^{\ell + 1} \ ,
    \label{equ:fusion}
\end{equation}

\noindent where $\uparrow$, $\odot$, and $\oplus$ represent the upsample, pixel-wise multiplication, and addition respectively. $\bm{F}^{\ell + 1}_{A}$ represents the updated feature after applying the attention maps. This simple adjustment constructs features based on coordinates, preserving robust constraints and establishing a foundation for subsequent multi-view correlation calculations.

\subsection{Aggregation}
\label{sec:aggregation}
With features extracted, homography warping is performed to transform the source view features into the reference perspective. Each source feature $\bm{F} \in \mathbb{R}^{G \times H \times W}$ is warped into $D$ fronto-parallel planes of the reference viewpoint to construct feature volumes $\{\bm{V}_i\}_{i=1}^N$. The $D$ depth hypotheses enable matching: the correct depth yields high feature correlation. The mathematical representation of the differentiable homography $\bm{H}_{i}(d)$ which warps each pixel at depth $d$ as:
\begin{equation}
    \bm{H}_{i}(d) = \bm{K}_{i} \cdot \bm{R}_{i} \cdot (\bm{I} - \frac{(\bm{t}_{0} - \bm{t}_{i}) \cdot \bm{n}_{0}^{\top}}{d}) \cdot \bm{R}_{0}^{\top} \cdot \bm{K}_{0}^{-1} \\
    \label{equ:homography}
\end{equation}

where $\{\bm{K}_{i},\bm{R}_{i},\bm{t}_{i}\}_{i=0}^{N}$ are the parameters that denote intrinsics, extrinsics and translations respectively, $\bm{n}_{0}$ represents the normal axis of reference viewpoints, $\bm{I}$ is the identity matrix. Then, the element-wise multiplication is applied to the reference feature volume $\bm{V}_{0}$ and each warped source feature volume $\{\bm{V}_{i}\}_{i=1}^{N}$ to obtain the pair-wise correlation: 
\begin{equation}
    \bm{Corr}_{i} = \bm{V}_{0} \odot \bm{V}_{i}
\end{equation}

Therefore, we get the 3D feature correlations $\bm{Corr} \in \mathbb{R}^{G \times D \times H \times W}$ between one pair of the reference image and the source image. Typically, those 3D feature correlations $\{\bm{Corr}_{i}\}_{i=1}^{N}$ are then aggregated into single cost volume $\bm{C} \in \mathbb{R}^{G \times D \times H \times W}$, which reflects the overall correlations of reference pixels and source pixels on the discrete depth planes. To further enhance the robustness and precision of the depth estimation, we integrate cross-view context into the cost volume. We use correlations from both the previous and current stages as guidance to extract contextual priors.

\noindent \textbf{\mdtwo(CVA).} In the first stage, we use the basic aggregation commonly employed in previous works~\cite{wang2022mvster,zhang2023geomvsnet}. The pair-wise attention weight $\bm{\mathcal{W}}$ is normalized by the number of channels and a temperature scaling factor $\epsilon$, and correlations are aggregated into one cost volume $\bm{C} \in \mathbb{R}^{G \times D \times H \times W}$:
\begin{equation}
        \bm{\mathcal{W}_{i}} = softmax \left( \frac{\sum_{g=1}^{G} \bm{Corr}_{i}[g]}{\epsilon} \right)
\end{equation}
\begin{equation}
        \bm{C} = \sum_{i=1}^{N} \frac{\bm{\mathcal{W}}_{i} \odot \bm{Corr}_{i}}{\sum_{i=1}^{N} \bm{\mathcal{W}}_{i}}
\end{equation}

Although this operation assigns a weight to each source view based on the sum of the feature values in that view, it fails to integrate low-level semantic information from the previous stage or capture the global connections between feature channels and depth assumptions across multiple views. Therefore, we design a lightweight module to extract cross-scale information in cross-view correlation.

As shown in Fig.~\ref{fig:architecture}(b), the feature channel $G$ and the depth channel $D$ of correlations are integrated to capture global correlation distribution information. Previous approaches such as GeoMVSNet~\cite{zhang2023geomvsnet} process these channels separately using additional 3D CNNs, neglecting the interdependencies between depth and feature dimensions. Besides, it incorporates geometry information by combining the probability volume and cost volume, which leads to computational overhead. In contrast, our method employs a 2D CNN to efficiently compress cross-scale contextual knowledge into a single channel. This approach significantly enhances efficiency.

We flatten the feature channels $G$ and depth hypotheses $D$ into a single dimension. This operation reformulates the correlation tensor into a shape of $(G \times D, H, W)$, where each flattened channel encodes combined information from both the feature and depth domains. This joint representation allows for the simultaneous processing of spatial and depth-aware features, enhancing the model's ability to capture multi-dimensional correlations. The contextual knowledge embedded in cross-view correlations extracted by the 2D CNN is concatenated with the original correlation to form a unified cost volume $\bm{C}$. This procedure can be described as:
\begin{equation}
    \bm{C}_{1} = Conv2D(\bm{C}^{\ell}), \bm{C}_{2} = Conv2D(\bm{C}^{\ell+1})
\end{equation}
\begin{equation}
    \bm{C}_{update} = Concat(\bm{C}, \bm{C}_{1}, \bm{C}_{2})
\end{equation}

Here, $\bm{C}^{\ell}$ represents the cost volume from the previous stage, $\bm{C}^{\ell+1}$ is the cost volume from the current stage, $Conv2D$ denotes a 2D convolution with batch normalization and ReLU activation, and $Concat$ refers to concatenation along the $G \times D$ dimension. The 2D CNN extracts information from the cost volume regarding connections between features and depths. The coarse branch encodes prior information, while the fine branch captures details from the current stage, enabling the model to capture contextual relationships across stages, depth assumptions, and feature channels.

\subsection{Regularization}
\label{sec:reg}

We employ a lightweight regularization network similar to~\cite{zhang2023geomvsnet}. The network consists of 3D CNNs that progressively downsample the input volumes, followed by transposed convolutions that upsample the volumes back to the original resolution. Skip connections are employed to combine features from corresponding layers during downsampling and upsampling. The final layer produces a probability volume $\bm{P} \in \mathbb{R}^{D \times H \times W}$ indicating confidence in each depth hypothesis.

Subsequently, we adopt a winner-takes-all strategy to derive the depth map. After applying softmax to $\bm{P}$, the index of the highest probability among the $D$ depth hypotheses for each pixel is selected. The corresponding depth value is then assigned as the pixel's depth. This depth value is used both to compute the loss for the current stage and as the center for defining depth hypotheses in the next stage.

\subsection{Loss Functions}
\label{sec:loss}

The loss functions are similar to those in previous methods~\cite{wang2022mvster,zhang2023geomvsnet}. Specifically, for the probability volume $\bm{P}$, a softmax operation is applied along the depth dimension $D$ to obtain a distribution of probabilities for each pixel. Each depth hypothesis corresponds to a specific depth value, and the classification is based on selecting the hypothesis with the highest probability. Ground truth labels are encoded using a one-hot scheme, where the depth hypothesis closest to the true depth value is marked as $1$, and all others are set to $0$. For simplicity, we use only the pixel-wise cross-entropy term, without additional components:
\begin{equation}
    \mathit{Loss}_\mathit{pw} = \sum_{z \in \Psi} (-P_{gt}(z) \ log[P(z)] ) \ ,
\end{equation}

\noindent where $\Psi$ represents the set of valid pixels with ground truth, $P$ denotes the estimated probability for each specific depth, while $P_{gt}$ corresponds to the probability volume of the ground truth. 
The overall loss is a weighted sum of $\mathit{Loss}_\mathit{pw}$ in Eqn.~\ref{equ:total-loss}, where $\lambda^\ell = 1$ among each stage in our experiments.
\begin{equation}
    \mathit{Loss} = \sum_{\ell=0}^L \lambda^\ell \ \mathit{Loss}_{pw} \ .
    \label{equ:total-loss}
\end{equation}
\section{Experiments}
\label{sec:experiment}

\subsection{Dataset}
\noindent \textbf{DTU}~\cite{dtu} is an indoor dataset comprising 124 distinct objects, with data for each scene meticulously captured from 49 different viewpoints under various lighting conditions. 

\noindent \textbf{Tanks and Temples (T$\&$T)}~\cite{tanksandtemples} is a challenging and realistic dataset, which presents a valuable resource for evaluating MVS methods in demanding real-world scenarios.

\noindent \textbf{BlendedMVS}~\cite{blendedmvs} is a large-scale synthetic training dataset with $17,000+$ images and precise ground truth 3D structures.

\subsection{Implementation Details}
Following the established paradigm, our model is trained on the DTU training set~\cite{dtu} and evaluated on the DTU testing set, using the same data split and view selection criteria as~\cite{yao2018mvsnet,gu2020cascade,wang2022mvster,zhang2023geomvsnet} for comparability. Additionally, we fine-tune our model on BlendedMVS~\cite{blendedmvs} and evaluate on T$\&$T~\cite{tanksandtemples}.

\noindent \textbf{Training.} We configure the number of input images to be $N=5$ for the DTU~\cite{dtu}, each with a resolution of $640 \times 512$ pixels. For the BlendedMVS~\cite{blendedmvs}, we employ a total of $N=7$ images, each with a resolution of $768 \times 576$ pixels. 
We train the model with an initial learning rate of $0.001$ for $15$ epochs on one NVIDIA RTX 4090 GPU, with a batch size of $4$.

\noindent \textbf{Evaluation.} For the DTU~\cite{dtu} evaluation, the images are cropped to a resolution of $1600 \times 1152$, and the number of input views remains $N=5$. When it comes to the T$\&$T~\cite{tanksandtemples}, we resize the height of the images to $1024$, while the width is retained at either $1920$ or $2048$, depending on the specific scene under evaluation. In this case, following the previous method, we increase the number of input views to $N=11$. For depth estimation on the DTU~\cite{dtu}, the inference runtime is 0.13 seconds and the model utilizes 3.21 GB of memory. On the T$\&$T~\cite{tanksandtemples} dataset, it executes within 0.6 seconds and consumes 7.37 GB of memory. For depth fusion, we adopt a dynamic fusion strategy~\cite{yan2020dense}, similar to previous approaches, to achieve dense point cloud integration.

\begin{figure*}[htbp!]
    \centering
    \includegraphics[width=0.98\linewidth]{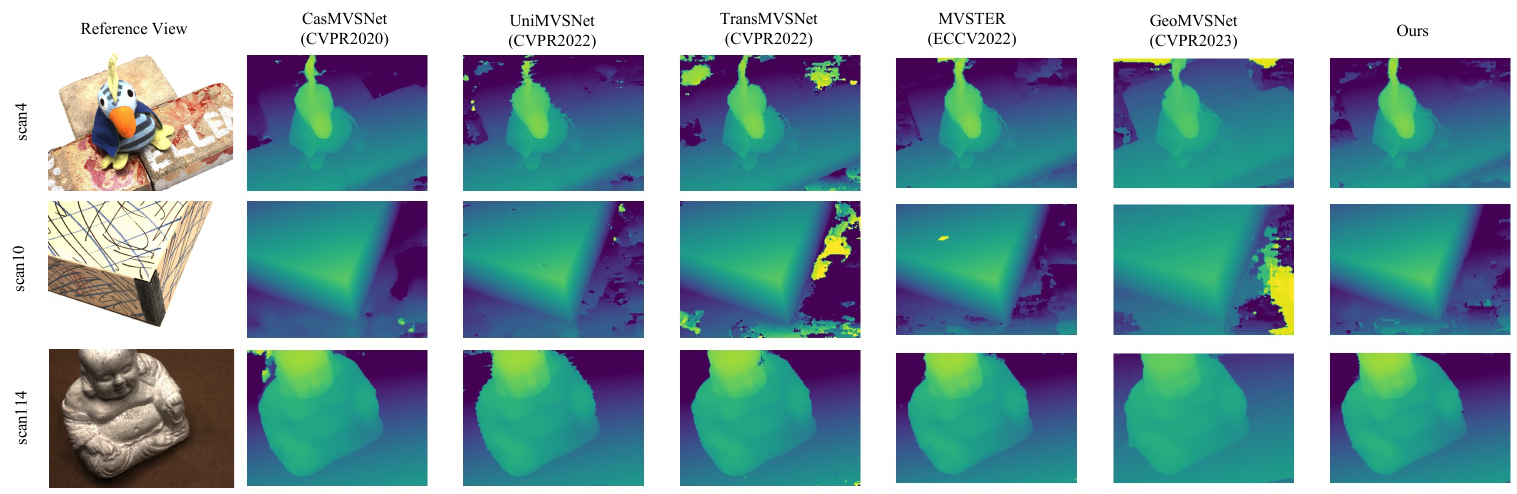}
    \vspace{-10pt}
    \caption{\textbf{Qualitative comparison with other methods on the DTU~\cite{dtu} dataset.} The depth map estimated by our method has a more complete and continuous surface and also has clearer outlines at the edges.}
    \label{fig:dtu-visual}
\vspace{-10pt}
\end{figure*}

\noindent \textbf{Metrics.} For the DTU~\cite{dtu}, we use distance metrics of point clouds to evaluate accuracy and completeness. Accuracy (Acc.) measures the distance from the reconstructed 3D points to the closest points in the ground truth, while completeness (Comp.) measures the distance from the ground truth to the reconstruction. To demonstrate that our method's advancements are due to the increased accuracy of the depth map, rather than fusion tricks, we also directly assess the depth map errors of various methods. For T$\&$T~\cite{tanksandtemples}, we adopt percentage-based metrics for accuracy and completeness. We use the official online evaluation platform for standardized assessments. 
For qualitative results, we used the official code and pre-trained weights to generate depth maps for visualization~Fig.~\ref{fig:dtu-visual} and error analysis~(Table~\ref{tab:depth-comparison}). Quantitative scores were taken from the original papers, except for memory and inference speed, which we re-evaluated under identical hardware for fairness.

\begin{table}[htb]
\renewcommand\arraystretch{1.15}
\centering
\vspace{-5pt}
\caption{\textbf{Quantitative comparison on DTU~\cite{dtu}.} $\ast$ means MVSTER~\cite{wang2022mvster} is trained on full-resolution images. The colors indicate rankings, with red representing the top position, orange indicating second place, and yellow marking third.}
\label{tab:DTU-comparison}
\resizebox{0.98\linewidth}{!}{
\begin{tabular}{ll|ccc|c|c}
    \hline
	& Method &  Acc.$\downarrow$ (\textit{mm}) & Comp.$\downarrow$ (\textit{mm}) & \textbf{Overall$\downarrow$(\textit{mm})} & \textbf{Time$\downarrow$(\textit{s})}& \textbf{GPU$\downarrow$(\textit{GB})}\\
	\hline
    & Gipuma\cite{galliani2015massively} & \cellcolor{softred}{0.283} & 0.873 & 0.578 & - & - \\
	& COLMAP\cite{colmap} & 0.400 & 0.664 & 0.532 & - & - \\
	\hline
    \multirow{11}{*}{\rotatebox{90}{\textbf{learning-based}}} 
	& R-MVSNet \cite{yao2019recurrent} & 0.383 & 0.452 & 0.417 & - & - \\
	& CasMVSNet \cite{gu2020cascade} & 0.325 & 0.385 & 0.355 & 0.49 & 5.4 \\
	& CVP-MVSNet \cite{yang2020cost} & \cellcolor{softorange}{0.296} & 0.406 & 0.351 & - & - \\
        & PatchmatchNet \cite{wang2021patchmatchnet} & 0.427 & 0.277 & 0.352 & 0.25 & \cellcolor{softred}{2.9} \\
	& EPP-MVSNet \cite{ma2021epp} & 0.413 & 0.296 & 0.355 & 0.52 & 8.2 \\
	& CDS-MVSNet \cite{giang2021curvature} & 0.352 & 0.280 & 0.316 & 0.40 & 4.5 \\
	& NP-CVP-MVSNet \cite{yang2022non} & 0.356 & 0.275 & 0.315 & 1.20 & 6.0\\
	& UniMVSNet \cite{peng2022rethinking} & 0.352 & 0.278 & 0.315 & 0.29 & \cellcolor{softyellow}{3.3} \\
	& TransMVSNet \cite{ding2022transmvsnet} & \cellcolor{softyellow}{0.321} & 0.289 & 0.305 & 0.99 & 3.8\\
	& MVSTER$^\ast$ \cite{wang2022mvster} & 0.340 & \cellcolor{softyellow}{0.266} & \cellcolor{softyellow}{0.303} & \cellcolor{softorange}{0.17} & 4.5 \\
        & GeoMVSNet \cite{zhang2023geomvsnet} & 0.331 & \cellcolor{softorange}{0.259} & \cellcolor{softorange}{0.295} & \cellcolor{softyellow}{0.26} & 5.9 \\
        \hline
	& Ours & 0.327 & \cellcolor{softred}{0.251} & \cellcolor{softred}{0.289} & \cellcolor{softred}{0.13} & \cellcolor{softorange}{3.2}\\
	\hline
\end{tabular}
}
\vspace{-10pt}
\end{table}

\begin{table}[htb]
    \centering
    \vspace{-5pt}
    \caption{\textbf{Depth map errors on DTU~\cite{dtu}.} The $ade$ represents the average absolute depth error (mm), while $tde(X)$ indicates the percentage of pixels with an error above X mm. The colors indicate rankings, with red representing the top position, orange indicating second place, and yellow marking third.}
    \label{tab:depth-comparison}
    \resizebox{0.98\linewidth}{!}{
    \begin{tabular}{l|cccccc}
        \hline
        Method & ade $\downarrow$ & tde(1) $\downarrow$ & tde(2) $\downarrow$ & tde(4) $\downarrow$ & tde(8) $\downarrow$ & tde(16) $\downarrow$ \\
        \hline
        MVSNet \cite{yao2018mvsnet} & 14.7356 & 28.22 & 20.01 & 16.19 & 14.00 & 12.18 \\
        CasMVSNet \cite{gu2020cascade} & \cellcolor{softyellow}{8.4086} & 25.41 & 17.80 & 14.06 & 11.42 & 8.97 \\
        CVP-MVSNet \cite{yang2020cost} & \cellcolor{softorange}{6.9875} & 26.75 & 18.80 & 14.10 & \cellcolor{softred}{10.22} & \cellcolor{softred}{6.75}\\
        TransMVSNet \cite{ding2022transmvsnet} & 15.1320 & 26.09 & 18.03 & 15.17 & 13.18 & 11.57 \\
        MVSTER \cite{wang2022mvster} & 9.3494 & 26.94 & 19.36 & 15.84 & 13.41 & 11.03 \\
        UniMVSNet \cite{peng2022rethinking} & 11.5872 & 28.90 & 19.66 & 13.73 & \cellcolor{softyellow}{11.02} & \cellcolor{softyellow}{8.92} \\
        GeoMVSNet \cite{zhang2023geomvsnet} & 11.6586 & \cellcolor{softred}{24.06} & \cellcolor{softred}{16.53} & \cellcolor{softorange}{13.28} & 11.08 & 9.21  \\
        
        \hline

        Ours & \cellcolor{softred}{6.6450} & \cellcolor{softorange}{24.38} & \cellcolor{softorange}{16.81} & \cellcolor{softred}{13.20} & \cellcolor{softorange}{10.55} & \cellcolor{softorange}{8.05} \\
        
        \hline
    \end{tabular}
    }
\vspace{-10pt}
\end{table}

\begin{table*}[htb]
    \centering
    \vspace{-5pt}
    \caption{\textbf{Quantitative comparison on Tanks and Temples~\cite{tanksandtemples}.} The colors indicate rankings, with red representing the top position, orange indicating second place, and yellow marking third.}
    \label{tab:TNT-comparison}
    \resizebox{0.98\linewidth}{!}{
    \begin{tabular}{c|ccccccccc}
    \hline 
    \multirow{2}{*}{\text {Method}} & \multicolumn{9}{c}{\text{Scene}} \\
    \cline{2-10}
    & \text { \textbf{Mean$\uparrow$} } & \text { Family } & \text { Francis } & \text { Horse } & \text { L.H. } & \text { M60 } & \text { Panther } & \text { P.G. } & \text { Train } \\
    \hline 
    \text { COLMAP\cite{colmap} }  & 42.14 & 50.41 & 22.25 & 25.63 & 56.43 & 44.83 & 46.97 & 48.53 & 42.04 \\

    \text { CasMVSNet\cite{gu2020cascade} }  & 56.42 & 76.36 & 58.45 & 46.20 & 55.53 & 56.11 & 54.02 & 58.17 & 46.56 \\
    
    \text { PatchmatchNet\cite{wang2021patchmatchnet} }  & 53.15 & 66.99 & 52.64 & 43.24 & 54.87 & 52.87 & 49.54 & 54.21 & 50.81 \\
    
    \text { UniMVSNet\cite{peng2022rethinking} }  & \cellcolor{softyellow}{64.36} & 81.20 & \cellcolor{softyellow}{66.43} & 53.11 & \cellcolor{softyellow}{63.46} & \cellcolor{softred}{66.09} & \cellcolor{softorange}{64.84} & \cellcolor{softyellow}{62.23} & 57.53 \\
    
    \text { TransMVSNet\cite{ding2022transmvsnet} }  & 63.52 & \cellcolor{softyellow}{80.92} & 65.83 & \cellcolor{softred}{56.94} & 62.54 & 63.06 & 60.00 & 60.20 & \cellcolor{softorange}{58.67} \\
    
    \text { MVSTER\cite{wang2022mvster} }  & 60.92 & 80.21 & 63.51 & 52.30 & 61.38 & 61.47 & 58.16 & 58.98 & 51.38 \\

    \text {GeoMVSNet\cite{zhang2023geomvsnet}}  & \cellcolor{softred}{65.89} & \cellcolor{softorange}{81.64} & \cellcolor{softorange}{67.53} & 55.78 & \cellcolor{softred}{68.02} & \cellcolor{softorange}{65.49} & \cellcolor{softred}{67.19} & \cellcolor{softred}{63.27} & \cellcolor{softyellow}{58.22} \\
    
    \hline

    \text {Ours}  & \cellcolor{softorange}{65.53} & \cellcolor{softred}{81.73} & \cellcolor{softred}{68.92} & \cellcolor{softorange}{56.59} & \cellcolor{softorange}{66.10} & \cellcolor{softyellow}{64.86} & \cellcolor{softyellow}{64.41} & \cellcolor{softorange}{62.33} & \cellcolor{softred}{59.26} \\
    
    \hline
    \end{tabular}
    }
\vspace{-10pt}
\end{table*}

\begin{table}[htb]
\renewcommand\arraystretch{1.15}
\centering
\vspace{-5pt}
\caption{\textbf{Ablations for proposed modules.} The model is evaluated on DTU~\cite{dtu}. Both modules can improve the accuracy and completeness of the point cloud.}
\label{tab:ablation-module}
\resizebox{0.98\linewidth}{!}{

\begin{tabular}{l|cc|ccc}
\hline
{Method} & Sec. \ref{sec:feature} & Sec. \ref{sec:aggregation} & Acc.$\downarrow$ (\textit{mm}) & Comp.$\downarrow$ (\textit{mm}) & \textbf{Overall$\downarrow$ (\textit{mm})} \\

\cline{2-5}

\hline
                          
baseline &   & & 0.350 & 0.276 & 0.313 \\ 

\hline

+ intra view  & \checkmark &  & 0.333 & 0.257 & 0.295 \\

+ cross view  &  & \checkmark & 0.327 & 0.255 & 0.291 \\

\hline

proposed & \checkmark & \checkmark & 0.327 & 0.251 & 0.289 \\ 

\hline
\end{tabular}

}
\vspace{-10pt}
\end{table}

\subsection{Benchmark Performance}
\noindent \textbf{DTU.} We compare our results with traditional methods and learning-based methods. We selected some representative visualizations, as shown in Fig.~\ref{fig:dtu-visual}. The depth map estimated by our method has a more complete and continuous surface and also has clearer outlines at the edges. Therefore, we can obtain accurate and complete point clouds, especially for the structures of the subject, even without rich textures such as the bird's head and statue edges. Using the official codes for point cloud evaluation, we present results in Tab.~\ref{tab:DTU-comparison}. Our method ranks first among the methods evaluated. Methods with similar scores need more inference time and GPU resources, while those with similar resource use perform worse.

Point cloud fusion involves various tricks to optimize benchmark results, but this contradicts the essence of MVS. To demonstrate that our method produces better depth maps rather than achieving high scores through parameter tuning, we analyze the estimated depth maps directly. We calculate the mean absolute error (mm) and the percentage of pixels with errors greater than $1mm$, $2mm$, etc., as shown in Tab.~\ref{tab:depth-comparison}. Our depth maps have lower average errors and more pixels within specific error thresholds. While GeoMVSNet~\cite{zhang2023geomvsnet} slightly outperforms us at the $1mm$ and $2mm$ thresholds, it performs worse at other levels and has larger absolute depth errors, indicating potential reliance on point cloud fusion to eliminate outliers. This explains why their depth maps are either highly accurate or significantly off, yet achieve strong point cloud scores.

\noindent \textbf{Tanks and Temples.} We further validate the generalization capability of our method on the T$\&$T~\cite{tanksandtemples} and report quantitative results in Tab.~\ref{tab:TNT-comparison}. 
Overall, our method achieves comparable precision and recall to GeoMVSNet~\cite{zhang2023geomvsnet}, but with the added benefits of reduced GPU memory usage—GeoMVSNet~\cite{zhang2023geomvsnet} needs 8.85 GB, whereas our method only requires 7.37 GB.

\subsection{Ablation Study}

Tab.~\ref{tab:ablation-module} and Tab.~\ref{tab:ablation-channel} show the ablation results of our method. The baseline MVSTER~\cite{wang2022mvster} method is re-implemented without the monocular depth estimator.

\noindent \textbf{\mdonenospace.} As shown in Tab.~\ref{tab:ablation-module}, \mdone which encodes both long-range dependencies and feature channel relationships can significantly improve the accuracy and completeness. This is because our method effectively captures coordinate information along with precise positional details, which is crucial for accurate understanding.

\noindent \textbf{\mdtwonospace.} As shown in Tab.~\ref{tab:ablation-module}, the cost volume aggregation equipped with the cross-view correlation derived from the previous stage can effectively improve the score because the correlations offer relationships to describe the positions of pixels, which means that some adjacent pixels are implicitly constrained to lie on a plane, this especially helps in texture-less regions. Besides, the contextual information extracted from correlations highlights the effect of various viewpoints, which enhances the robustness of the method to handle unreliable inputs.

\noindent \textbf{The number of channels in cross-view aggregation.} As shown in Tab.~\ref{tab:ablation-channel}, the cross-view correlation between the current stage and the previous stage can provide additional information to enhance performance. We conducted ablations by varying the number of channels used for correlation, testing up to $4$ channels within the bounds of the original volume. We found that more channels did not significantly improve completeness but slightly decreased accuracy, resulting in a minor overall score reduction. This suggests that while correlation data captures essential trends across viewpoints, additional channels may introduce noise without significant benefit, and increase memory usage. Therefore, we optimized our model by limiting the channels to $1$.

\begin{table}[htb]
\renewcommand\arraystretch{1.15}
\centering
\vspace{-5pt}
\caption{\textbf{Ablations for the number of channels of cross-view aggregation.} The model is evaluated on DTU~\cite{dtu}. The $\text{Num}_{\text{c}}$ is the number of channels from the previous stage and the $\text{Num}_{\text{f}}$ is the number of channels from the current stage.}
\label{tab:ablation-channel}
\resizebox{0.98\linewidth}{!}{

\begin{tabular}{l|cc|ccc}
\hline
{Method} & $\text{Num}_{\text{c}}$ & $\text{Num}_{\text{f}}$ & Acc.$\downarrow$ (\textit{mm}) & Comp.$\downarrow$ (\textit{mm}) & \textbf{Overall$\downarrow$ (\textit{mm})} \\

\cline{2-5}

\hline
                          
baseline+intra-view  & 0 & 0 & 0.333 & 0.257 & 0.295 \\ 

\hline
   & 0 & 1 & 0.3271 & 0.2585 & 0.2928 \\

   & 1 & 0 & 0.3300 & 0.2582 & 0.2941 \\

\hline

   & 2 & 2 & 0.3296 & 0.2492 & 0.2894 \\

   & 3 & 3 & 0.3323 & 0.2503 & 0.2913 \\

   & 4 & 4 & 0.3280 & 0.2544 & 0.2912 \\

\hline

proposed & 1 & 1 & 0.3270 & 0.2510 & 0.2890 \\ 

\hline
\end{tabular}

}
\vspace{-10pt}
\end{table}
\section{Conclusion}
\label{sec:conclusion}
In this paper, we propose \ours which explicitly integrates intra-view and cross-view relationships for depth estimation. Specifically, we construct \mdonenospace, which leverages the positional knowledge within a single image, enhancing robust cost matching without incorporating complicated external dependencies. Besides, we introduce a lightweight \mdtwo that efficiently utilizes the cross-scale contextual information from correlations. The proposed method is extensively evaluated on public datasets, consistently achieving competitive performance against the state-of-the-arts, while requiring no extra input and lower computational resources. Our method still needs large labeled datasets to train the model. Future work could focus on developing unsupervised methods to overcome these limitations. 

\bibliographystyle{IEEEbib}
\bibliography{icme2025references}

\begin{thebibliography}{10}

\bibitem{gu2020cascade}
Xiaodong Gu, Zhiwen Fan, et~al.,
\newblock ``Cascade cost volume for high-resolution multi-view stereo and stereo matching,''
\newblock in {\em Proceedings of the IEEE/CVF Conference on Computer Vision and Pattern Recognition}, 2020, pp. 2495--2504.

\bibitem{cheng2020deep}
Shuo Cheng, Zexiang Xu, et~al.,
\newblock ``Deep stereo using adaptive thin volume representation with uncertainty awareness,''
\newblock in {\em Proceedings of the IEEE/CVF Conference on Computer Vision and Pattern Recognition}, 2020, pp. 2524--2534.

\bibitem{yang2020cost}
Jiayu Yang, Wei Mao, et~al.,
\newblock ``Cost volume pyramid based depth inference for multi-view stereo,''
\newblock in {\em Proceedings of the IEEE/CVF Conference on Computer Vision and Pattern Recognition}, 2020, pp. 4877--4886.

\bibitem{etmvsnet}
Tianqi Liu, Xinyi Ye, et~al.,
\newblock ``When epipolar constraint meets non-local operators in multi-view stereo,''
\newblock in {\em Proceedings of the IEEE/CVF International Conference on Computer Vision (ICCV)}, October 2023, pp. 18088--18097.

\bibitem{ding2022transmvsnet}
Yikang Ding, Wentao Yuan, et~al.,
\newblock ``Transmvsnet: Global context-aware multi-view stereo network with transformers,''
\newblock in {\em Proceedings of the IEEE/CVF Conference on Computer Vision and Pattern Recognition}, 2022, pp. 8585--8594.

\bibitem{wang2022mvster}
Xiaofeng Wang, Zheng Zhu, et~al.,
\newblock ``Mvster: Epipolar transformer for efficient multi-view stereo,''
\newblock in {\em European Conference on Computer Vision}. Springer, 2022, pp. 573--591.

\bibitem{cao2022mvsformer}
Chenjie Cao, Xinlin Ren, et~al.,
\newblock ``Mvsformer: Multi-view stereo by learning robust image features and temperature-based depth,''
\newblock {\em Transactions of Machine Learning Research}, 2023.

\bibitem{cao2024mvsformer++}
Xinlin~Ren Chenjie~Cao et~al.,
\newblock ``Mvsformer++: Revealing the devil in transformer's details for multi-view stereo,''
\newblock in {\em International Conference on Learning Representations (ICLR)}, 2024.

\bibitem{zhang2023geomvsnet}
Zhe Zhang, Rui Peng, et~al.,
\newblock ``Geomvsnet: Learning multi-view stereo with geometry perception,''
\newblock in {\em Proceedings of the IEEE/CVF Conference on Computer Vision and Pattern Recognition}, 2023, pp. 21508--21518.

\bibitem{wu2024gomvs}
Jiang Wu, Rui Li, et~al.,
\newblock ``Gomvs: Geometrically consistent cost aggregation for multi-view stereo,''
\newblock in {\em CVPR}, 2024.

\bibitem{lin2017feature}
Tsung-Yi Lin, Piotr Doll{\'a}r, et~al.,
\newblock ``Feature pyramid networks for object detection,''
\newblock in {\em Proceedings of the IEEE conference on computer vision and pattern recognition}, 2017, pp. 2117--2125.

\bibitem{dtu}
Rasmus Jensen, Anders Dahl, et~al.,
\newblock ``Large scale multi-view stereopsis evaluation,''
\newblock in {\em Proceedings of the IEEE conference on computer vision and pattern recognition}, 2014, pp. 406--413.

\bibitem{guo2019group}
Xiaoyang Guo, Kai Yang, et~al.,
\newblock ``Group-wise correlation stereo network,''
\newblock in {\em Proceedings of the IEEE/CVF Conference on Computer Vision and Pattern Recognition}, 2019, pp. 3273--3282.

\bibitem{yao2018mvsnet}
Yao Yao, Zixin Luo, et~al.,
\newblock ``Mvsnet: Depth inference for unstructured multi-view stereo,''
\newblock in {\em Proceedings of the European conference on computer vision (ECCV)}, 2018, pp. 767--783.

\bibitem{yao2019recurrent}
Yao Yao, Zixin Luo, et~al.,
\newblock ``Recurrent mvsnet for high-resolution multi-view stereo depth inference,''
\newblock in {\em Proceedings of the IEEE/CVF conference on computer vision and pattern recognition}, 2019, pp. 5525--5534.

\bibitem{tanksandtemples}
Arno Knapitsch, Jaesik Park, et~al.,
\newblock ``Tanks and temples: Benchmarking large-scale scene reconstruction,''
\newblock {\em ACM Transactions on Graphics}, vol. 36, no. 4, 2017.

\bibitem{blendedmvs}
Yao Yao, Zixin Luo, et~al.,
\newblock ``Blendedmvs: A large-scale dataset for generalized multi-view stereo networks,''
\newblock in {\em Proceedings of the IEEE/CVF Conference on Computer Vision and Pattern Recognition}, 2020, pp. 1790--1799.

\bibitem{yan2020dense}
Jianfeng Yan, Zizhuang Wei, et~al.,
\newblock ``Dense hybrid recurrent multi-view stereo net with dynamic consistency checking,''
\newblock in {\em European conference on computer vision}. Springer, 2020, pp. 674--689.

\bibitem{galliani2015massively}
Silvano Galliani, Katrin Lasinger, et~al.,
\newblock ``Massively parallel multiview stereopsis by surface normal diffusion,''
\newblock in {\em Proceedings of the IEEE International Conference on Computer Vision}, 2015, pp. 873--881.

\bibitem{colmap}
Johannes~L Schonberger and Jan-Michael Frahm,
\newblock ``Structure-from-motion revisited,''
\newblock in {\em Proceedings of the IEEE conference on computer vision and pattern recognition}, 2016, pp. 4104--4113.

\bibitem{wang2021patchmatchnet}
Fangjinhua Wang, Silvano Galliani, et~al.,
\newblock ``Patchmatchnet: Learned multi-view patchmatch stereo,''
\newblock in {\em Proceedings of the IEEE/CVF Conference on Computer Vision and Pattern Recognition}, 2021, pp. 14194--14203.

\bibitem{ma2021epp}
Xinjun Ma, Yue Gong, et~al.,
\newblock ``Epp-mvsnet: Epipolar-assembling based depth prediction for multi-view stereo,''
\newblock in {\em Proceedings of the IEEE/CVF International Conference on Computer Vision}, 2021, pp. 5732--5740.

\bibitem{giang2021curvature}
Khang~Truong Giang, Soohwan Song, et~al.,
\newblock ``Curvature-guided dynamic scale networks for multi-view stereo,''
\newblock {\em arXiv:2112.05999}, 2021.

\bibitem{yang2022non}
Jiayu Yang, Jose~M Alvarez, et~al.,
\newblock ``Non-parametric depth distribution modelling based depth inference for multi-view stereo,''
\newblock in {\em Proceedings of the IEEE/CVF Conference on Computer Vision and Pattern Recognition}, 2022, pp. 8626--8634.

\bibitem{peng2022rethinking}
Rui Peng, Rongjie Wang, et~al.,
\newblock ``Rethinking depth estimation for multi-view stereo: A unified representation,''
\newblock in {\em Proceedings of the IEEE/CVF Conference on Computer Vision and Pattern Recognition}, 2022, pp. 8645--8654.

\end{thebibliography}


\clearpage
\setcounter{page}{1}

\section{Supplementary Material}
In the supplementary material, we present more details that are not included in the main text, including:

\begin{itemize}
    \item Depth map fusion method which integrates depth maps into 3D point clouds.
    \item Extended comparison with additional recent works based on reviewer suggestions.
    \item Depth map visualizations: comparisons between our method and other approaches.
    \item Point cloud visualizations: comparisons with other methods and our final reconstructed point clouds on DTU~\cite{dtu} and Tanks and Temples~\cite{tanksandtemples}.
\end{itemize}

\subsection{Depth Map Fusion}
\label{sec:fusion}
In previous Multi-View Stereo (MVS) approaches, various fusion techniques have been utilized to integrate predicted depth maps from multiple viewpoints into a coherent point cloud. In this study, we adopt the dynamic checking strategy proposed in~\cite{yan2020dense} for depth filtering and fusion.

On the DTU~\cite{dtu} dataset, we filter the confidence map of the final stage using a confidence threshold to assess photometric consistency. For geometric consistency, we apply the following criteria:

\begin{equation}
\label{eq:dypcd}
\text{err}_c < \text{thresh}_c, \quad \text{err}_d < \text{thresh}_d
\end{equation}

Here, \(\text{err}_c\) and \(\text{err}_d\) represent the reprojection coordinate error and relative error of reprojection depth, respectively. \(\text{thresh}_c\) and \(\text{thresh}_d\) indicate the corresponding thresholds.

On the Tanks and Temples\cite{tanksandtemples} benchmark, we adjust hyperparameters for each scene following the approach outlined in~\cite{peng2022rethinking}, including confidence thresholds and geometric criteria. Our model is fine-tuned on the BlendedMVS dataset~\cite{blendedmvs} for reconstructing scenes in this benchmark.


\subsection{Depth Map Visualizations}
\label{sec:depthmap}
\begin{figure*}[htbp!]
    \centering
    \includegraphics[width=0.98\linewidth]{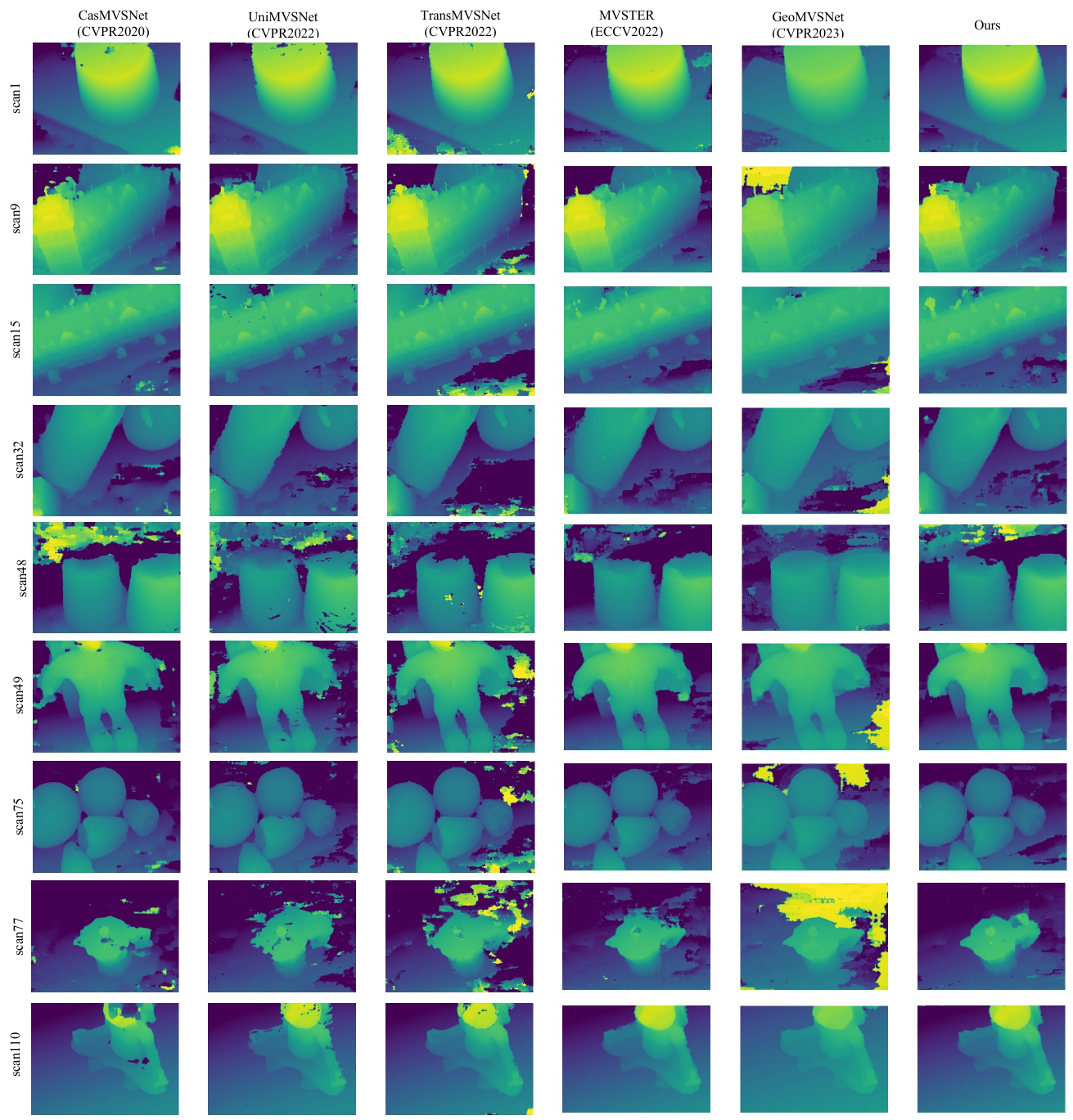}
    \caption{\textbf{Qualitative comparison with other methods on DTU~\cite{dtu}.} The depth map estimated by our method has a more complete and continuous surface and also has clearer outlines at the edges.}
    \label{fig:dtu-visual-supp}
\end{figure*}

The depth map visualizations illustrate the performance of our method compared to other approaches. These visualizations (Fig.~\ref{fig:dtu-visual-supp}) highlight the accuracy and completeness of the predicted depth maps, showcasing improvements in fine details and handling of challenging regions such as edges and textureless areas.

\subsection{Extended Comparison with Additional Recent Works}
Our focus is on uncovering the core essence of the problem so we prefer not to rely on pre-trained models or additional inputs as priors, which is why we specifically compare with works that train from scratch to ensure a fair evaluation. Nonetheless, we include the quantitative results of CT-MVSNet~(MMM2024), MVSFormer++~(ICLR2024), and GoMVS~(CVPR2024) on the DTU dataset in Table~\ref{tab:DTU-comparison-supp} as requested, to provide additional comparisons and references for potential readers. Overall, our proposed method achieves competitive performance against SOTA while requiring no extra input and lower computational resources. 
(1) \textbf{Our method consistently outperforms CT-MVSNet.} Our method achieves significantly better accuracy, completeness, and overall performance. 
(2) \textbf{GoMVS relies on additional normal inputs.} In particular, it requires normal map estimation as a preprocessing step, which imposes a strong planar constraint that undoubtedly improves its benchmark performance. However, this also means that the results are highly dependent on the accuracy of the normal estimation method. 
Moreover, even if we disregard the preprocessing time required to obtain normal maps, GoMVS still consumes more than twice the memory and takes over three times longer to run compared to the proposed method.
(3) \textbf{MVSFormer++ relies heavily on a pre-trained model.} While its overall score is better, it is crucial to note that MVSFormer++ benefits from extensive engineering and experimental optimizations. It not only leverages a pre-trained DINOv2 model for feature extraction but also incorporates efficiency-focused components like FlashAttention. However, these optimizations come at the cost of significantly increased model complexity—MVSFormer++ has over 30 times~(783MB) more parameters than our model~(20MB). Moreover, its training is highly resource-intensive, requiring four A6000 GPUs~(48GB each) for a full day, whereas our model can be trained in less than a day using just a single RTX 4090. In addition, MVSFormer++ demands more GPU memory and longer inference time, as shown in Table~\ref{tab:DTU-comparison-supp}, making it less efficient in practical deployment.

\begin{table}[htb]
\centering
\caption{\textbf{Quantitative comparison on DTU.} The point cloud metrics are taken from the respective papers. All runtime and memory usage data are measured on the same hardware (RTX 4090) for a fair comparison. Note that CT-MVSNet does not provide a pre-trained model, without which we could not test on our machine.}
\label{tab:DTU-comparison-supp}
\resizebox{1\linewidth}{!}{
\renewcommand{\arraystretch}{1.5}
\begin{tabular}{l|ccc|c|c}
    \hline
	Method &  Acc.$\downarrow$ (\textit{mm}) & Comp.$\downarrow$ (\textit{mm}) & \textbf{Overall$\downarrow$(\textit{mm})} & \textbf{Time$\downarrow$(\textit{s})}& \textbf{GPU$\downarrow$(\textit{GB})}\\
    \hline
    \multirow{3}{*}{\rotatebox{90}{\textbf{}}} 
        CT-MVSNet & 0.341 & 0.264 & 0.302 & - & - \\
        MVSFormer++ & 0.309 & 0.252 & 0.281 & 0.30 & 5.6 \\
        GoMVS & 0.347 & 0.227 & 0.287 & 0.45 & 7.7 \\
	\hline
	Ours & 0.327 & 0.251 & 0.289 & 0.13 & 3.2\\
	\hline
\end{tabular}
}
\end{table}

\subsection{Point Cloud Visualizations}
\label{sec:pointcloud}
\begin{figure*}[htbp!]
    \centering
    \includegraphics[width=0.98\linewidth]{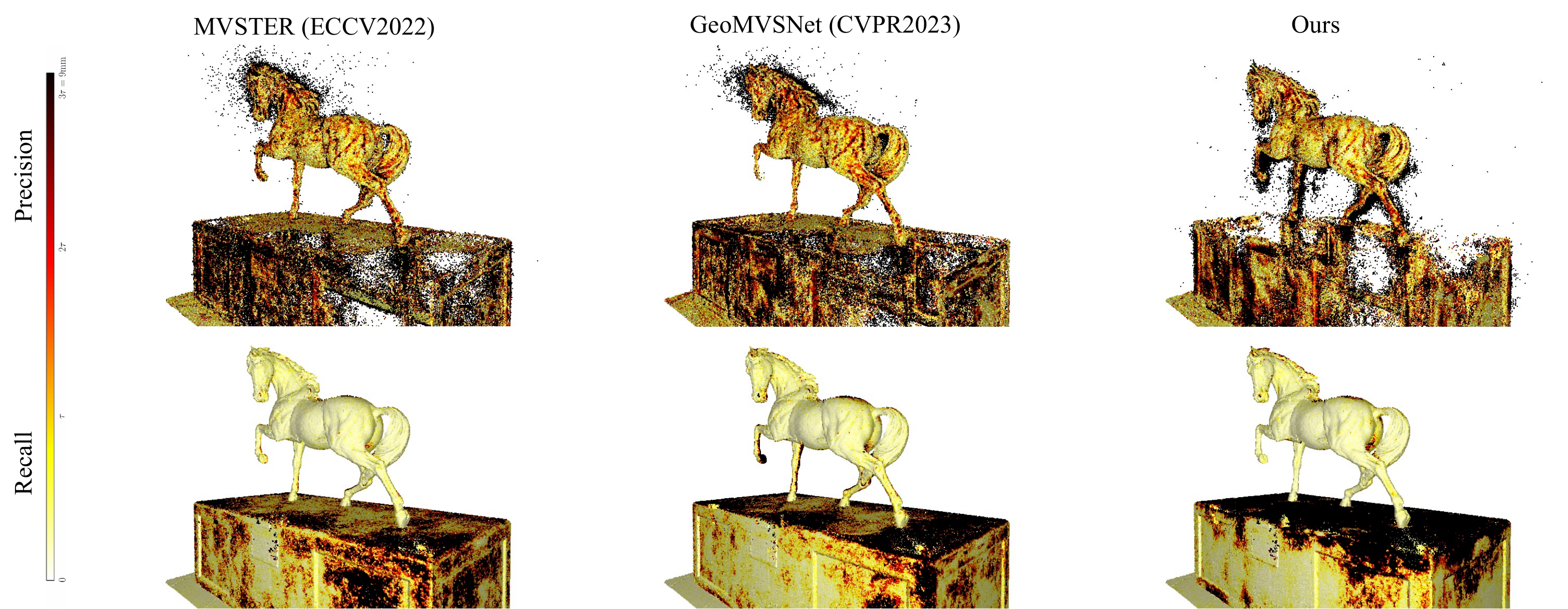}
    \caption{\textbf{Point clouds error comparison of state-of-the-art methods on the Tanks and Temples dataset~\cite{tanksandtemples}.} $\tau$ is the scene-relevant distance threshold determined officially and darker means large error. The first row shows \textit{Precision} and the second row shows \textit{Recall}. Taking the \textit{Horse} in the intermediate subset as an example, our method is able to reduce large amounts of outliers while ensuring completeness.}
    \vspace{-1ex}
    \label{fig:tnt-error}
\end{figure*}

Fig.~\ref{fig:tnt-error} illustrates the error comparison of the point clouds. Taking the \textit{Horse} in Tanks and Temples~\cite{tanksandtemples} dataset as an example, our method is able to reduce large amounts of outliers while ensuring completeness.

\begin{figure*}[htbp!]
    \centering
    \includegraphics[width=0.98\linewidth]{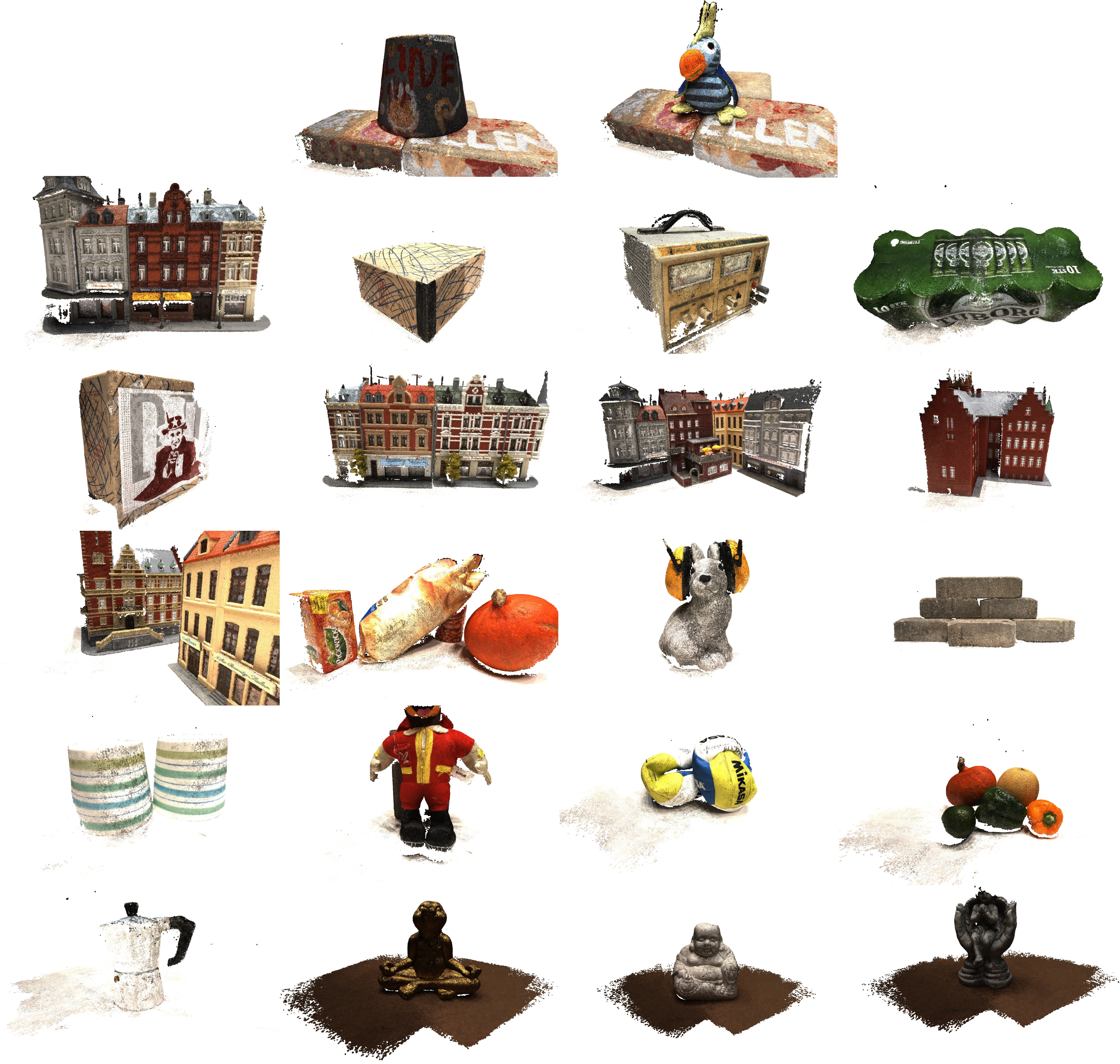}
    \caption{\textbf{Point clouds reconstructed for all scenes from DTU~\cite{dtu} dataset.}}
    \vspace{-1ex}
    \label{fig:dtu-pointcloud}
\end{figure*}

\begin{figure*}[htbp!]
    \centering
    \includegraphics[width=0.98\linewidth]{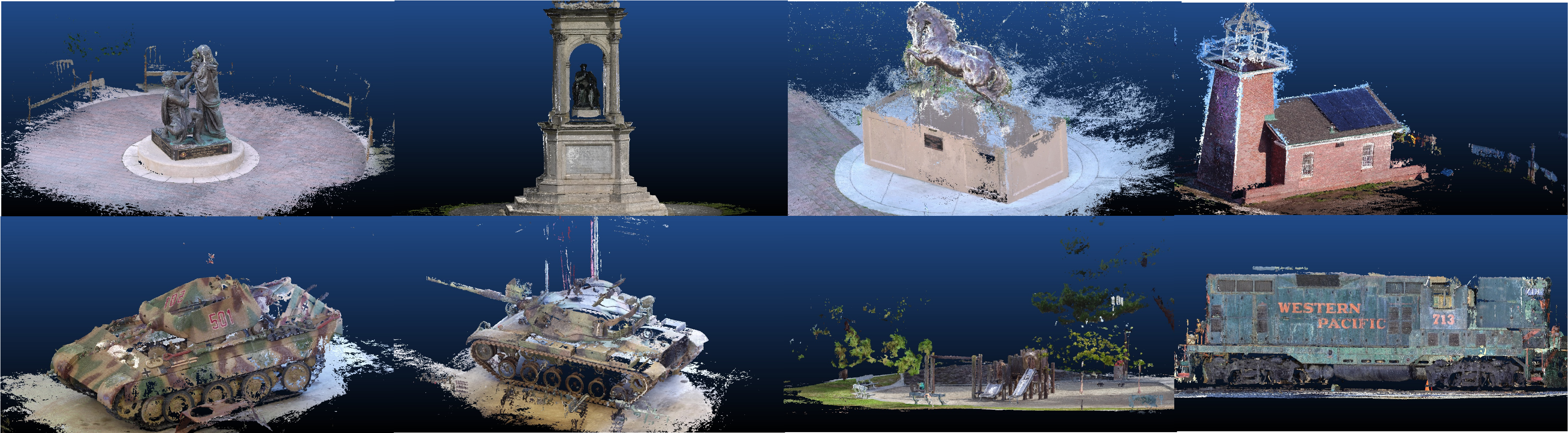}
    \caption{\textbf{Point clouds reconstructed for all scenes from Tanks and Temples~\cite{tanksandtemples} dataset.}}
    \vspace{-1ex}
    \label{fig:tnt-pointcloud}
\end{figure*}

Besides, we show all points clouds reconstructed using our method on DTU~\cite{dtu} dataset and Tanks and Temples~\cite{tanksandtemples} dataset, respectively. As illustrated in Fig.~\ref{fig:dtu-pointcloud} and Fig.~\ref{fig:tnt-pointcloud}, our method demonstrates robust reconstruction capabilities across various scales, effectively handling both small objects and large-scale scenes. 

\end{document}